%% file: neurips_2026_arxiv_preprint.tex
\documentclass{article}

\usepackage[preprint]{neurips_2026}

\usepackage[utf8]{inputenc}
\usepackage[T1]{fontenc}
\usepackage{hyperref}
\usepackage{url}
\usepackage{booktabs}
\usepackage{amsfonts}
\usepackage{amssymb}
\usepackage{amsmath}
\usepackage{nicefrac}
\usepackage{microtype}
\usepackage{xcolor}
\usepackage{graphicx}
\graphicspath{{../figures/publication-ready/}}
\usepackage{algorithm}
\usepackage{algorithmic}
\usepackage{multirow}
\usepackage{subcaption}

\newcommand{\parents}{\mathrm{Pa}}
\newcommand{\BIC}{\mathrm{BIC}}
\newcommand{\RSS}{\mathrm{RSS}}
\newcommand{\MSE}{\mathrm{MSE}}

\usepackage{acro}
\DeclareAcronym{dag}{short=DAG, long=directed acyclic graph}
\DeclareAcronym{bn}{short=BN, long=Bayesian network}
\DeclareAcronym{bnp}{short=BNP, long=Bayesian nonparametric}
\DeclareAcronym{bge}{short=BGe, long=Bayesian Gaussian equivalent}
\DeclareAcronym{mse}{short=MSE, long=mean squared error}
\DeclareAcronym{bic}{short=BIC, long=Bayesian information criterion}
\DeclareAcronym{atp}{short=ATP, long=American Trends Panel}
\DeclareAcronym{em}{short=EM, long=expectation-maximization}
\DeclareAcronym{dp}{short=DP, long=Dirichlet process}
\DeclareAcronym{ari}{short=ARI, long=adjusted Rand index}
\DeclareAcronym{nmi}{short=NMI, long=normalized mutual information}
\DeclareAcronym{shd}{short=SHD, long=structural Hamming distance}
\DeclareAcronym{mcmc}{short=MCMC, long=Markov chain Monte Carlo}
\DeclareAcronym{rmse}{short=RMSE, long=root mean squared error}

\title{Heterogeneous Ordinal Structure Learning with\\Bayesian Nonparametric Complexity Discovery}

\author{%
  Amir Rafe, Ph.D.\\
  Texas State University\\
  San Marcos, USA\\
  \texttt{amir.rafe@txstate.edu}\\
  ORCID: \href{https://orcid.org/0000-0002-4089-2088}{0000-0002-4089-2088}
  \And
  Subasish Das, Ph.D.\\
  Texas State University\\
  San Marcos, USA\\
  \texttt{subasish@txstate.edu}\\
  ORCID: \href{https://orcid.org/0000-0002-1671-2753}{0000-0002-1671-2753}
}

\begin{document}

\maketitle

\begin{abstract}
Public attitudes toward artificial intelligence are heterogeneous, ordinally measured, and poorly captured by any single dependency graph. Existing ordinal structure learners assume a shared \ac{dag} across all respondents; recent heterogeneous ordinal graphical-model approaches focus on subgroup discovery rather than confirmatory cluster-specific \ac{dag} estimation; and latent profile analyses discard dependency structure entirely. We introduce a heterogeneous ordinal structure-learning framework combining monotone Gaussian score embedding, \ac{bnp} complexity discovery via a truncated stick-breaking prior, and confirmatory fixed-$K$ estimation with cluster-specific sparse \acs{dag} learning. The key methodological insight is a discovery-to-confirmation workflow: the nonparametric stage calibrates plausible archetype complexity, while inner-validated confirmatory refitting yields stable, interpretable structural estimates. On the 2024 Pew American Trends Panel AI attitudes survey, Wave 152 (W152) survey, ($N{=}4{,}788$, 8~ordinal items), the confirmatory $K^*{=}5$ model reduces holdout transformed-score \ac{mse} by 25.8\% over a single-graph baseline and by 4.6\% over mixture-only clustering. A controlled tiered semi-synthetic benchmark calibrated to W152 structure validates recovery across difficulty regimes and transparently reveals failure modes under stress conditions.
\end{abstract}

\acresetall

\section{Introduction}
\label{sec:intro}

Public attitudes toward artificial intelligence are increasingly measured through large-scale ordinal survey batteries~\citep{pewai2024, zhang2019artificial, scantamburlo2024ai}. Standard analyses typically make two simplifying assumptions: the population shares one dependency structure, and ordinal responses can be treated as continuous without distortion. Both are questionable. If subpopulations differ in how trust, regulation, and perceived benefits interact, a single shared graph mischaracterizes every group. This concern is substantive as well as statistical. Recent measurement work argues that AI attitudes contain separable positive and negative dimensions, with trust-related beliefs and institutional confidence varying in ways that do not collapse cleanly to a single pro-versus-anti axis~\citep{schepman2020gaais, schepman2023gaais, zhang2019artificial, scantamburlo2024ai}. In exactly this setting, averaging respondents into one population graph risks obscuring the heterogeneity that survey researchers care about most.

Adjacent methodological literatures each address only part of this problem. Ordinal-aware \ac{bn} methods respect category ordering but learn one shared graph~\citep{luo2021ordinal, grzegorczyk2024ordinal, ni2025ordinal}. Mixed graphical and psychometric network models estimate dependence among mixed or ordinal variables, but are typically undirected and fitted at the population level~\citep{lee2015mixed, epskamp2018networks}. Model-based clustering and latent profile analysis support subgroup discovery in respondent-level distributions, yet focus on profile recovery rather than cluster-specific dependency structure~\citep{fraley2002model, shum2024latent}. Mixture-of-\acs{dag} models allow subgroup-specific structure but usually operate on continuous or categorical observations~\citep{thiesson1998score, saeed2020identifiability, castelletti2023heterogeneous, castelletti2024bnpmixture}. Recent heterogeneous ordinal or categorical graphical-model work moves closer to the present setting~\citep{wang2025heterogeneousordinal, ferrini2026graphical}. What remains useful here is a stable workflow for learning heterogeneous ordinal structure and reporting it defensibly.

We therefore study a three-stage framework: (i)~a monotone Gaussian score embedding for ordinal data (\S\ref{sec:embedding}), (ii)~\acs{bnp} complexity discovery via a truncated stick-breaking \ac{dp} mixture~\citep{ferguson1973bayesian, sethuraman1994constructive} (\S\ref{sec:bnp}), and (iii)~confirmatory fixed-$K$ heterogeneous structure estimation with cluster-specific sparse \acp{dag} (\S\ref{sec:confirmatory}). The core methodological idea is \emph{discovery to confirmation}: use the nonparametric stage to learn plausible complexity, then report an inner-validated fixed-dimensional estimator rather than either arbitrary segmentation or the raw nonparametric fit. This design is more defensible for empirical reporting because the discovery stage calibrates complexity while the confirmatory stage yields a stable model for comparison and substantive interpretation.

We organize the empirical study around four research questions. \textbf{RQ1} (Complexity Discovery): How many latent archetypes does the data support, and does nonparametric discovery agree with inner-validated model selection? \textbf{RQ2} (Archetype Structure): Do the discovered archetypes exhibit qualitatively distinct dependency graphs, or do they differ only in mean profiles? \textbf{RQ3} (Model Comparison): Does the heterogeneous ordinal model improve holdout transformed-score prediction relative to single-graph and mixture-only baselines? \textbf{RQ4} (Validation): Can the framework recover known structure in controlled semi-synthetic data, and is it robust to prior specification and item-set perturbation? We evaluate the method on the 2024 Pew \ac{atp} W152, a nationally representative U.S.\ survey on AI attitudes, using an analytic sample of $N{=}4{,}788$ respondents and 8 ordinal items.

The paper makes three contributions. First, it extends ordinal structure learning to subgroup-specific sparse \acp{dag} through a heterogeneous workflow combining monotone score embedding and cluster-specific graphs. Second, it introduces a discovery-to-confirmation strategy that uses the \acs{bnp} stage to calibrate complexity and an inner-validated fixed-$K$ refit for reporting. Third, it shows on Pew W152 and a controlled benchmark that the approach recovers interpretable archetypes, improves fit over strong baselines, and makes its stability limits explicit.

\section{Method}
\label{sec:method}

\begin{figure}[t]
  \centering
  \includegraphics[width=\linewidth]{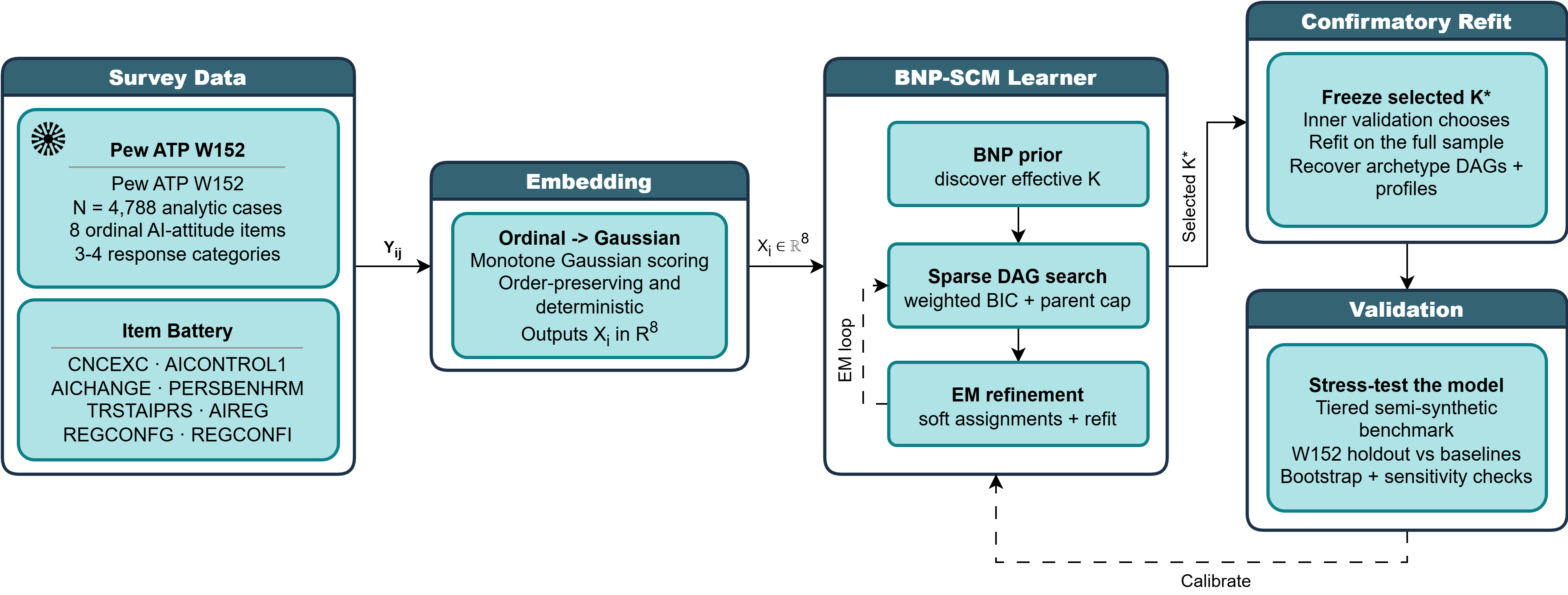}
  \caption{\textbf{Discovery-to-confirmation pipeline.} Ordinal survey responses are embedded via monotone Gaussian scoring, then a \acs{bnp} mixture discovers plausible archetype complexity. Inner validation selects $K^*$, and a confirmatory refit produces stable archetype \acp{dag} and profiles for substantive reporting.}
  \label{fig:pipeline}
\end{figure}

We present a heterogeneous ordinal structure-learning framework whose three-stage pipeline, ordinal embedding (\S\ref{sec:embedding}), \acs{bnp} complexity discovery (\S\ref{sec:bnp}), and confirmatory fixed-$K$ estimation (\S\ref{sec:confirmatory}), is built on a cluster-specific structural model described in \S\ref{sec:structural}. Figure~\ref{fig:pipeline} provides an overview.

\subsection{Monotone Gaussian Score Embedding}
\label{sec:embedding}

Let respondent $i = 1, \ldots, N$ answer $J$ ordinal items with $Y_{ij} \in \{1, \ldots, C_j\}$. Ordinal responses cannot be treated as continuous without distorting dependency estimation. We embed each item into a monotone Gaussian score that preserves category ordering while enabling sparse Gaussian \acs{dag} estimation.

For item $j$, let $p_{jc} = P(Y_j = c)$ denote the empirical category mass. Define the cumulative midpoint:
\begin{equation}
  u_{jc} = \sum_{\ell < c} p_{j\ell} + \tfrac{1}{2}\, p_{jc},
  \label{eq:midpoint}
\end{equation}
and the category score:
\begin{equation}
  s_j(c) = \Phi^{-1}(u_{jc}),
  \label{eq:score}
\end{equation}
where $\Phi^{-1}$ is the standard normal quantile function. Each respondent's observed ordinal value is then replaced by its category score:
\begin{equation}
  X_{ij} = s_j(Y_{ij}),
  \label{eq:transform}
\end{equation}
and all subsequent structure learning operates on the transformed matrix $\mathbf{X} \in \mathbb{R}^{N \times J}$.

This embedding is monotone ($c_1 < c_2 \Rightarrow s_j(c_1) < s_j(c_2)$), deterministic (unlike ordinal probit data augmentation~\citep{albert1993bayesian}), and produces approximately standard-normal marginals. It preserves Spearman rank correlations exactly and avoids the computational cost of \ac{mcmc}-based latent-variable models while respecting ordinal structure.

\subsection{Archetype-Specific Structural Model}
\label{sec:structural}

Let $z_i \in \{1, \ldots, K\}$ denote respondent $i$'s latent archetype. Conditional on archetype $k$, the transformed item vector $\mathbf{X}_i = (X_{i1}, \ldots, X_{iJ})^\top$ follows a sparse linear-Gaussian \acs{dag}:
\begin{equation}
  X_{ij} = \beta_{k0,j} + \sum_{m \in \parents_k(j)} \beta_{km \to j}\, X_{im} + \varepsilon_{ijk}, \quad \varepsilon_{ijk} \sim \mathcal{N}(0,\, \sigma^2_{kj}),
  \label{eq:dag}
\end{equation}
where $G_k$ is the directed acyclic graph for archetype $k$, $\parents_k(j)$ is the parent set of node $j$ in $G_k$, $\beta_{km \to j}$ is the edge weight from node $m$ to node $j$, and $\sigma^2_{kj}$ is the residual variance. Crucially, each archetype has its own adjacency matrix, parent sets, regression coefficients, and residual variances.

\paragraph{Cluster-specific \ac{bic} scoring.}
For a given cluster $k$ with soft responsibilities $r_{ik}$, the local weighted \acs{bic} for node $j$ is:
\begin{equation}
  \BIC_{kj}(G_k) = n_k \log\!\left(\frac{\RSS_{kj}}{n_k}\right) + d_{kj} \log n_k,
  \label{eq:bic_node}
\end{equation}
where $n_k = \sum_i r_{ik}$ is the effective cluster mass, $\RSS_{kj} = \sum_i r_{ik}(X_{ij} - \hat{X}_{ij})^2$ is the responsibility-weighted residual sum of squares, and $d_{kj}$ is the number of regression coefficients including the intercept. The graph-level objective is:
\begin{equation}
  \BIC_k(G_k) = \sum_{j=1}^{J} \BIC_{kj}(G_k).
  \label{eq:bic_graph}
\end{equation}
Graph search uses greedy acyclicity-constrained edge operations (add, delete, reverse) under a maximum-parent cap~\citep{chickering2002optimal}.

\subsection{Bayesian Nonparametric Complexity Discovery}
\label{sec:bnp}

The number of archetypes $K$ should be learned from data rather than specified a priori. We use a truncated stick-breaking representation of a \acs{dp} mixture~\citep{sethuraman1994constructive}:
\begin{equation}
  V_k \sim \mathrm{Beta}(1, \alpha), \quad \pi_k = V_k \prod_{\ell < k}(1 - V_\ell), \quad k = 1, \ldots, K_{\max},
  \label{eq:stickbreak}
\end{equation}
where $\alpha > 0$ is the concentration parameter and $K_{\max}$ is a finite truncation level. This stage discovers plausible archetype complexity by fitting the full mixture-of-\acp{dag} model and observing how many components receive non-negligible mass.

\paragraph{Iterative soft-assignment refinement.}
The heterogeneous estimator alternates between responsibility updates (E-step) and responsibility-weighted \acs{dag} fitting (M-step)~\citep{dempster1977maximum}:

\textbf{E-step.} Update responsibilities using the current parameters:
\begin{equation}
  r_{ik}^{(t+1)} \propto \tilde{\pi}_k^{(t)} \cdot p(\mathbf{X}_i \mid G_k^{(t)}, \theta_k^{(t)}),
  \label{eq:estep}
\end{equation}
where the smoothed cluster weights are:
\begin{equation}
  \tilde{\pi}_k^{(t)} = \frac{\sum_i r_{ik}^{(t)} + \alpha / K}{N + \alpha},
  \label{eq:smooth_weights}
\end{equation}
followed by row normalization over $k$.

\textbf{M-step.} For each cluster $k$, refit the cluster-specific \acs{dag} using the updated responsibilities:
\begin{equation}
  (G_k^{(t+1)}, \theta_k^{(t+1)}) = \arg\max_{G, \theta} \sum_i r_{ik}^{(t+1)} \log p(\mathbf{X}_i \mid G, \theta) - \lambda\,\mathrm{pen}(G),
  \label{eq:mstep}
\end{equation}
where $\mathrm{pen}(G)$ is the \acs{bic} complexity term from Eq.~\ref{eq:bic_node} and $\lambda{=}1$ throughout. Because greedy \acs{dag} search does not guarantee a global optimum, this constitutes a generalized \ac{em} procedure. The refinement loop terminates when both the assignment-change rate and the objective change fall below a convergence threshold (see Appendix~\ref{app:convergence}).

\subsection{Confirmatory Fixed-$K$ Estimation}
\label{sec:confirmatory}

The \acs{bnp} discovery stage provides a data-driven estimate of archetype complexity, but the nonparametric fit may over-split in practice (see \S\ref{sec:synthetic}). For stable empirical reporting, we select the confirmatory complexity via inner validation:
\begin{equation}
  K^* = \arg\min_{K \in \mathcal{K}_{\mathrm{grid}}} \MSE_{\mathrm{val}}(K),
  \label{eq:kstar}
\end{equation}
where $\mathcal{K}_{\mathrm{grid}} = \{2, 3, 4, 5, 6\}$ and the transformed-score holdout error is:
\begin{equation}
  \MSE = \frac{1}{N_{\mathrm{test}} \cdot J} \sum_{i \in \mathrm{test}} \sum_{j=1}^{J} \left(X_{ij} - \hat{X}_{ij}\right)^2,
  \label{eq:mse}
\end{equation}
with $\hat{X}_{ij} = \sum_k r_{ik}\,\hat{X}_{ij\mid k}$ denoting the responsibility-weighted prediction from the cluster-specific \acp{dag}. All model comparisons in the paper use this shared transformed-score objective induced by the common monotone embedding.

After selecting $K^*$, we refit the heterogeneous model with exactly $K^*$ components on the full analytic sample. This confirmatory estimator is used for all substantive archetype reporting. The discovery-to-confirmation workflow constitutes the key methodological contribution: the \acs{bnp} stage discovers complexity, inner validation selects it, and the fixed-$K^*$ refit confirms it. The full procedure is summarized in Algorithm~\ref{alg:main}.

\begin{algorithm}[H]
\caption{Discovery-to-Confirmation Heterogeneous Ordinal Structure Learning}
\label{alg:main}
\begin{algorithmic}[1]
\REQUIRE Ordinal data $\{Y_{ij}\}_{i=1}^N$, grid $\mathcal{K}_{\mathrm{grid}}$, concentration $\alpha$, truncation $K_{\max}$
\STATE \textbf{Embed:} Compute $X_{ij} = s_j(Y_{ij})$ via Eqs.~(\ref{eq:midpoint})--(\ref{eq:transform})
\STATE \textbf{Discover:} Fit \acs{bnp} mixture-of-\acp{dag} with stick-breaking prior (Eq.~\ref{eq:stickbreak}), alternating E/M steps (Eqs.~\ref{eq:estep}--\ref{eq:mstep}). Record effective $\hat{K}_{\mathrm{BNP}}$.
\FOR{$K \in \mathcal{K}_{\mathrm{grid}}$}
  \STATE Split data into train/validation folds
  \STATE Fit fixed-$K$ heterogeneous model on train set
  \STATE Evaluate $\MSE_{\mathrm{val}}(K)$ on held-out fold (Eq.~\ref{eq:mse})
\ENDFOR
\STATE \textbf{Select:} $K^* = \arg\min_K \MSE_{\mathrm{val}}(K)$
\STATE \textbf{Confirm:} Refit fixed-$K^*$ model on full sample $\rightarrow$ archetype \acp{dag} $\{G_k\}$, profiles $\{\theta_k\}$, assignments $\{z_i\}$
\RETURN $K^*$ archetype \acp{dag}, profiles, and assignments
\end{algorithmic}
\end{algorithm}

\section{Experimental Setup}
\label{sec:setup}

We evaluate the framework on the 2024 Pew \ac{atp} 
W152~\citep{pewai2024}, a nationally representative U.S.\@ survey on public attitudes toward artificial intelligence. The full wave contains $N{=}5{,}410$ respondents; after complete-case restriction on the eight-item main battery our analytic sample is $N{=}4{,}788$ (88.5\% retention). We verified the data-processing pipeline by reproducing official Pew topline percentages, obtaining a mean absolute deviation of 0.34 percentage points on the full weighted wave and 0.73 percentage points on the analytic sample. The battery comprises eight ordinal indicators with between three and six observed response categories, spanning concern versus excitement toward AI (\texttt{CNCEXC}), perceived personal benefit versus harm (\texttt{PERSBENHRM}), expected impact on daily life (\texttt{AICHANGE}), AI control over important decisions (\texttt{AICONTROL1}), trust in AI for personal decisions (\texttt{TRSTAIPRS}), demand for regulation (\texttt{AIREG}), and confidence in government and industry regulation (\texttt{REGCONFG}, \texttt{REGCONFI}). Comparable AI-attitude batteries are sparse across earlier \ac{atp} waves, so we use W152 as the primary structural dataset and treat cross-wave evidence as descriptive context only (Appendix~\ref{app:crosswave}).

We compare four models, all sharing the same monotone Gaussian score embedding. The Single-Graph Baseline learns one \acs{dag} over the full sample via standard ordinal \acs{bn} scoring. The Fixed-$K{=}5$ Mixture Only model applies Gaussian mixture clustering without cluster-specific \acp{dag}. The \acs{bnp} Discovery model is the full nonparametric mixture-of-\acp{dag} from the discovery stage. The Fixed-$K{=}5$ Mixture + \acs{dag} model is our proposed confirmatory heterogeneous estimator. Holdout evaluation uses an 80/20 random split; inner validation for $K$-selection uses five-fold cross-validation within the training set.

Because real-data evaluation cannot verify cluster recovery in the absence of ground truth, we additionally construct a tiered semi-synthetic benchmark calibrated to W152 structure. The generator is drawn from the same latent-score family as the fitted model, so we interpret this benchmark as a controlled recovery diagnostic rather than a universal misspecification test. The benchmark defines four difficulty regimes: \emph{Easy} (high cluster separation, large graph differences), \emph{Moderate} (realistic parameters), \emph{Hard} (low separation, small graph differences, class imbalance), and \emph{Stress} (minimal signal, designed to reveal failure modes). Each tier uses three replications. We report predictive transformed-score \acs{mse}, \ac{ari}~\citep{hubert1985comparing}, \ac{nmi}~\citep{vinh2010information}, and cluster-specific \ac{shd}~\citep{tsamardinos2006max}.

\section{Results}
\label{sec:results}

\subsection{Complexity Discovery}
\label{sec:complexity}

\begin{figure}[t]
  \centering
  \includegraphics[width=0.95\linewidth]{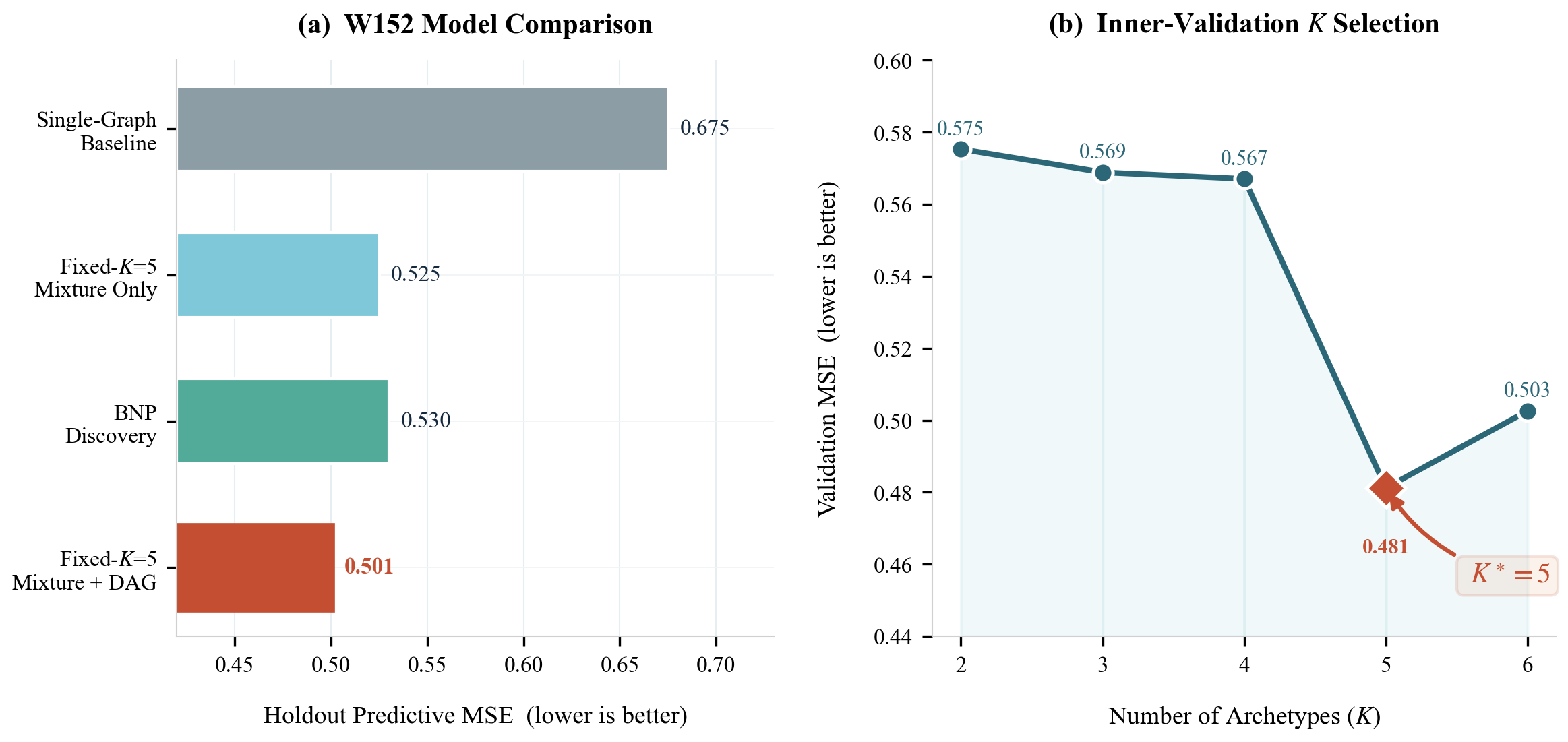}
  \caption{\textbf{W152 model comparison and $K$-selection.} (a)~Holdout \acs{mse} across four models; the confirmatory fixed-$K{=}5$ heterogeneous model achieves the lowest error. (b)~Inner-validation $K$-selection curve showing $K^*{=}5$ as the clear minimum.}
  \label{fig:overview}
\end{figure}

The \acs{bnp} discovery fit retains five nontrivial archetypes when initialized with $K_{\max}{=}10$ components. Inner validation over $K \in \{2, 3, 4, 5, 6\}$ independently selects $K^*{=}5$ with the lowest validation \acs{mse} of 0.481 (Figure~\ref{fig:overview}). The agreement between the nonparametric discovery and the inner-validation selection strengthens the $K{=}5$ choice, providing converging evidence from two complementary model-selection strategies.

\subsection{Archetype-Specific Dependency Structures}
\label{sec:archetypes}

\begin{figure}[t]
  \centering
  \includegraphics[width=\linewidth]{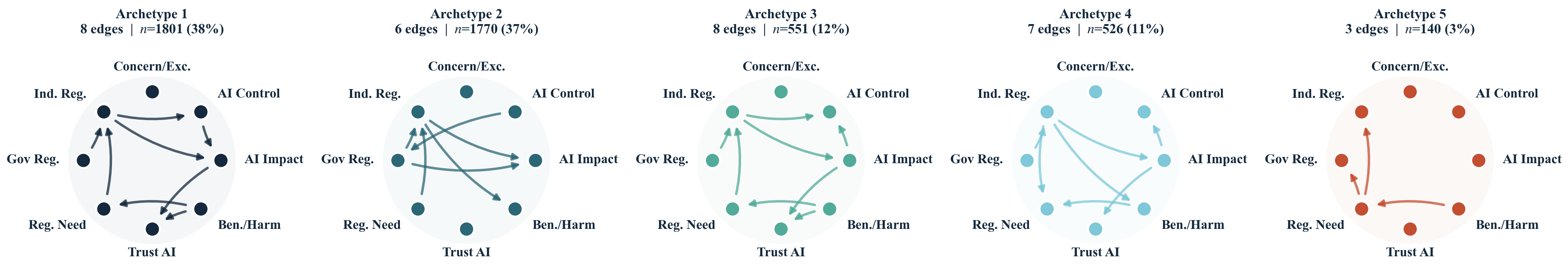}
  \caption{\textbf{Archetype-specific \acp{dag}.} Five cluster-specific \acp{dag} from the confirmatory $K^*{=}5$ fit. Edge patterns differ substantially across archetypes, particularly around regulation and trust items.}
  \label{fig:dags}
\end{figure}

The confirmatory $K^*{=}5$ fit reveals five archetypes with distinct dependency structures. Figure~\ref{fig:dags} displays the five cluster-specific \acp{dag}. The two largest archetypes (Arch.~1: $n{=}1{,}801$, 37.6\%; Arch.~2: $n{=}1{,}770$, 37.0\%) differ in both graph density (8 versus 6 edges) and edge configuration. Archetypes~3 and~4 ($n{=}551$ and $n{=}526$) capture regulatory-focused subgroups with distinct trust-regulation linkages. Archetype~5 ($n{=}140$, 2.9\%) exhibits minimal dependency structure (3 edges), consistent with extreme or decoupled attitude positions.

\begin{figure}[t]
  \centering
  \includegraphics[width=0.93\linewidth]{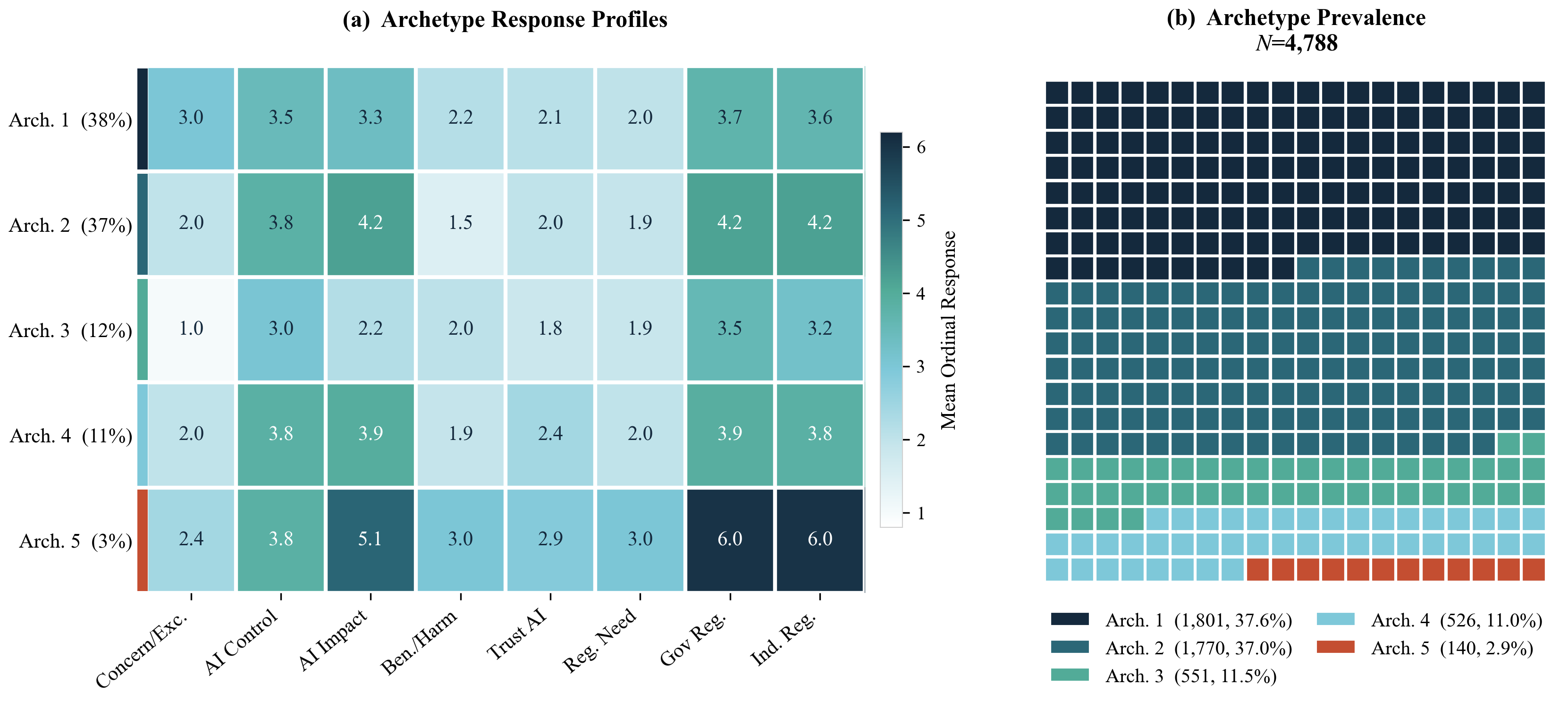}
  \caption{\textbf{Archetype anatomy.} (a)~Response profile heatmap showing mean ordinal scores per item per archetype. (b)~Prevalence waffle chart ($N{=}4{,}788$).}
  \label{fig:anatomy}
\end{figure}

Regulation items (\texttt{AIREG}, \texttt{REGCONFG}, \texttt{REGCONFI}) form a strongly connected subgraph in Archetypes~1, 3, and~4, but not in Archetype~2. Trust in AI (\texttt{TRSTAIPRS}) shows different parent and child patterns across archetypes, supporting the hypothesis that dependency structure, not just mean profiles, differs across subpopulations. Figure~\ref{fig:anatomy} provides a complementary view: the response profile heatmap reveals how mean ordinal scores vary across archetypes and items, while the prevalence waffle chart confirms the dominance of the two largest archetypes.

Taken together, Figures~\ref{fig:dags} and~\ref{fig:anatomy} suggest that heterogeneity is not merely a one-dimensional shift in response intensity. The two largest archetypes are similarly prevalent yet differ in both graph density and local organization, while the smaller regulatory-focused archetypes differ in trust-regulation linkage despite partially overlapping mean profiles. This pattern is precisely the setting in which a single shared graph or mean-only segmentation would compress substantively different response systems into an average that fits no subgroup especially well.

\subsection{Model Comparison}
\label{sec:comparison}

Table~\ref{tab:model_comparison} summarizes the holdout transformed-score comparison, from which three findings emerge. First, heterogeneity matters: both heterogeneous models substantially outperform the single-graph baseline (\acs{mse} of 0.530 and 0.501 versus 0.675). Second, cluster-specific structure adds value under this shared objective, as the mixture+\acs{dag} model (0.501) outperforms mixture-only clustering (0.525), yielding a 4.6\% relative \acs{mse} reduction. Third, confirmation beats raw discovery: the fixed-$K$ confirmatory fit (0.501) outperforms the \acs{bnp} discovery fit (0.530) on holdout prediction, validating the two-stage workflow.

The comparison against mixture-only clustering is especially informative because both models share the same ordinal embedding and the same fixed complexity. The remaining performance gap therefore cannot be attributed to a different preprocessing pipeline or to extra flexibility in the number of components. Instead, it isolates the contribution of cluster-specific dependency structure itself: after segmentation is held fixed, allowing different \acp{dag} within clusters still improves predictive fit.

\begin{table}[t]
  \caption{\textbf{W152 holdout model comparison.} All models share the same monotone Gaussian embedding and are evaluated on transformed-score \acs{mse}. The confirmatory fixed-$K{=}5$ heterogeneous model achieves the best predictive fit.}
  \label{tab:model_comparison}
  \centering
  \begin{tabular}{lcc}
    \toprule
    Model & \acs{mse} $\downarrow$ & $\Delta$ vs.\ Baseline \\
    \midrule
    Single-Graph Baseline        & 0.675 & --- \\
    \acs{bnp} Discovery          & 0.530 & $-$21.5\% \\
    Fixed-$K{=}5$ Mixture Only   & 0.525 & $-$22.2\% \\
    \textbf{Fixed-$K{=}5$ Mix.\ + \acs{dag}} & \textbf{0.501} & $\mathbf{-25.8\%}$ \\
    \bottomrule
  \end{tabular}
\end{table}

\subsection{Semi-Synthetic Benchmark}
\label{sec:synthetic}

\begin{figure}[t]
  \centering
  \includegraphics[width=0.87\linewidth]{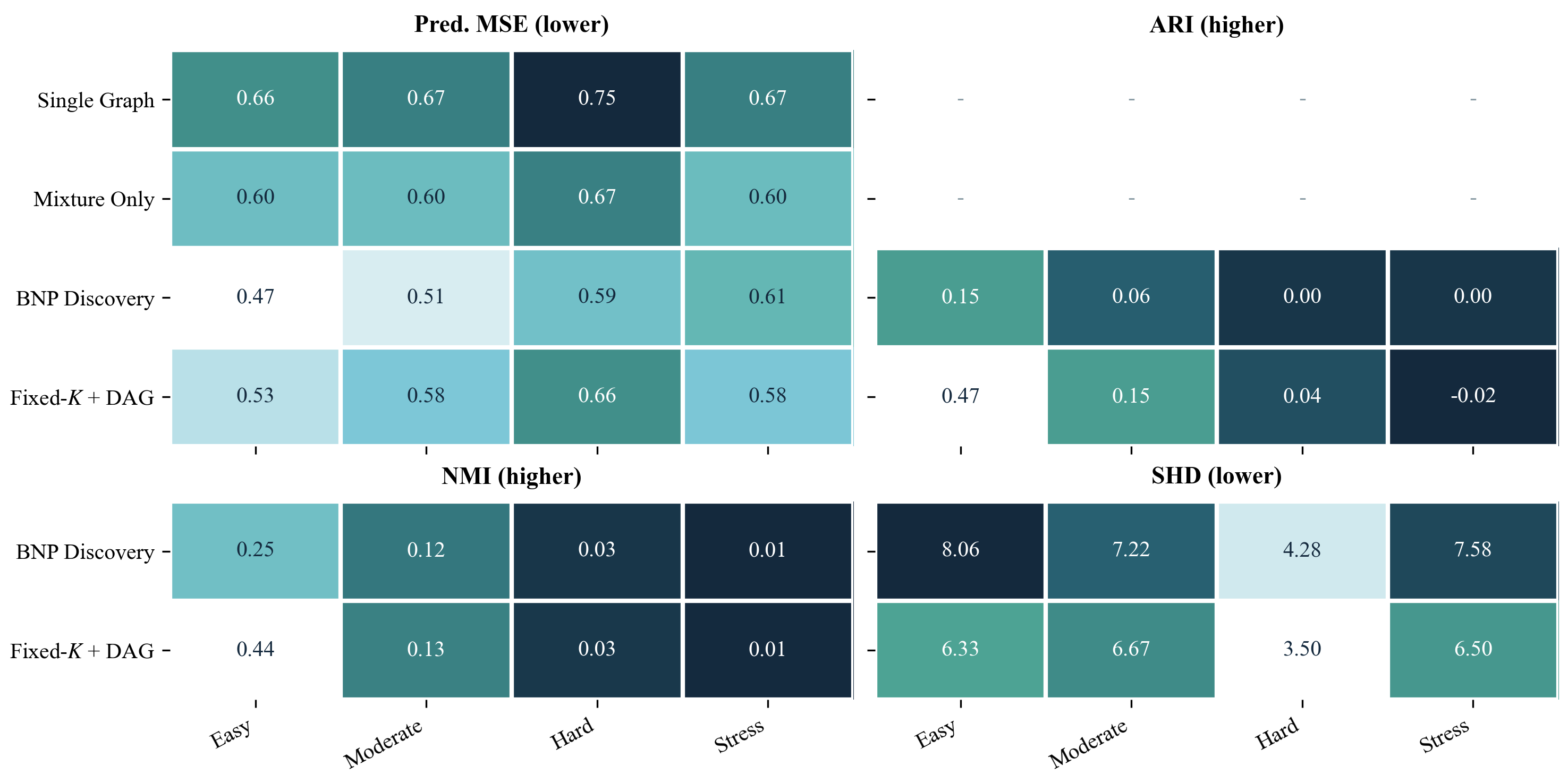}
  \caption{\textbf{Tiered semi-synthetic benchmark.} Predictive transformed-score \acs{mse}, \acs{ari}, \acs{nmi}, and \acs{shd} across four difficulty regimes, aggregated over three replications per tier. The fixed-$K$+\acs{dag} model excels on cluster and graph recovery; all methods fail honestly under Stress conditions.}
  \label{fig:synthetic}
\end{figure}

Figure~\ref{fig:synthetic} presents the tiered benchmark results across four difficulty regimes. Because the benchmark is generated from the same latent-score family as the fitted model, it should be read as controlled recovery evidence rather than an exhaustive misspecification test. Within that scope, heterogeneous models consistently outperform the single-graph baseline on prediction in all tiers. The fixed-$K$+\acs{dag} model achieves the best cluster recovery (\acs{ari}, \acs{nmi}) and graph recovery (\acs{shd}) in the recoverable regimes. The \acs{bnp} fit tends to over-split, particularly in easier settings, which supports the discovery-to-confirmation workflow. In the Stress tier, all methods produce near-zero \acs{ari}, providing honest failure-mode reporting and confirming that no method fabricates structure where none exists.

\subsection{Sensitivity Analysis}
\label{sec:sensitivity}

\begin{figure}[t]
  \centering
  \includegraphics[width=0.83\linewidth, height=0.6\linewidth]{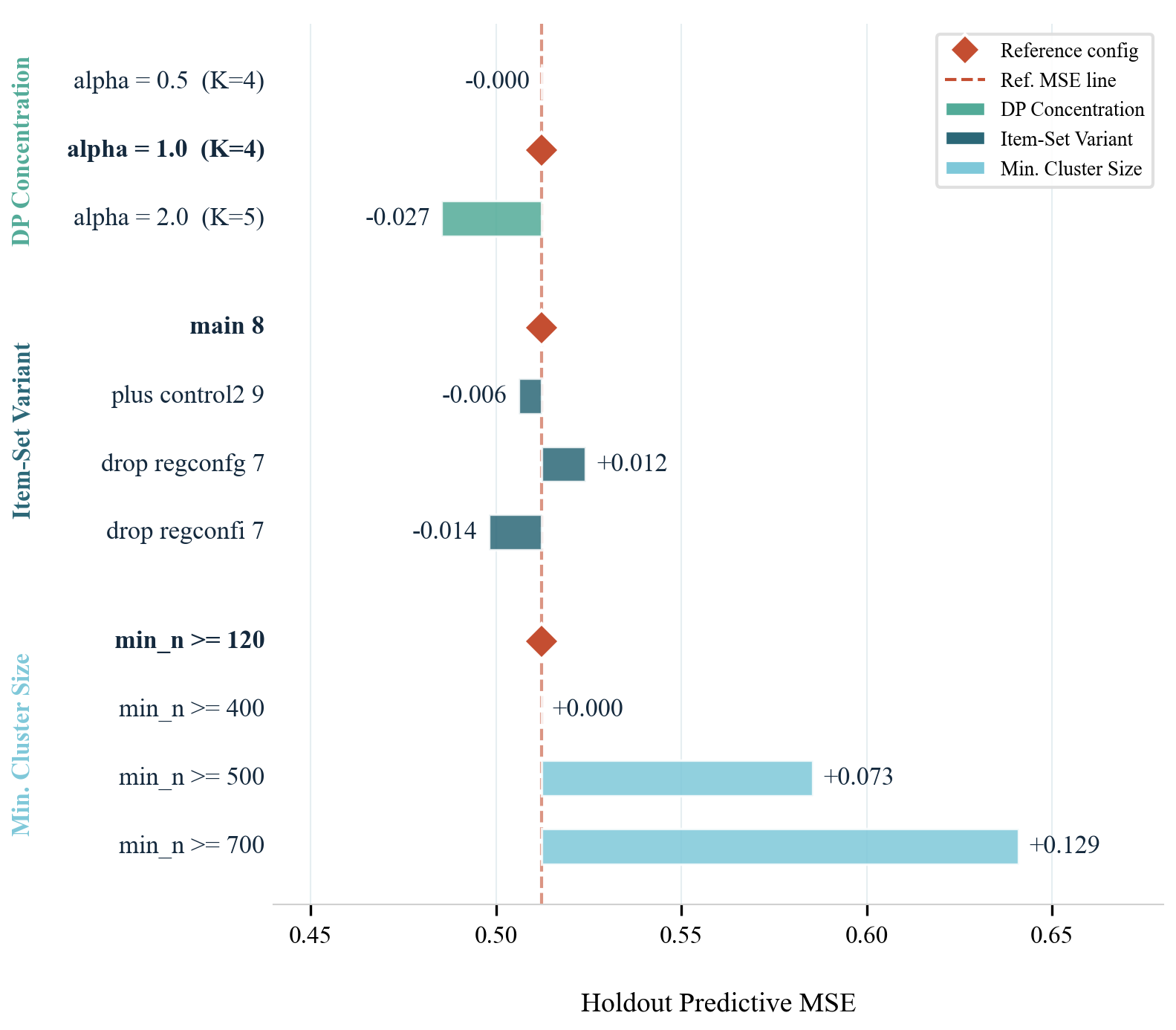}
  \caption{\textbf{Sensitivity analysis.} Forest plot showing \acs{mse} sensitivity to \acs{dp} concentration $\alpha$, item-set variants ($\pm 1$ item), and minimum cluster size. The model is robust across prior and item perturbations; forcing large minimum clusters degrades heterogeneous signal.}
  \label{fig:sensitivity}
\end{figure}

We assess robustness along three axes, summarized in Figure~\ref{fig:sensitivity}. Varying the \acs{dp} concentration $\alpha \in \{0.5, 1.0, 2.0\}$ yields stable \acs{mse} with changes below 0.03, indicating that the model is not sensitive to the prior on cluster proliferation. Item-set perturbation (adding or removing one item) produces \acs{mse} changes below 0.015, confirming that results are not driven by any single indicator. The model remains robust when the minimum cluster size is set up to $n_{\min}{=}400$; forcing $n_{\min}{\geq}500$ degrades \acs{mse} by $+$0.073, indicating that small but genuine archetypes contribute meaningful predictive signal.

Bootstrap stability analysis over 20 resamples yields a mean assignment agreement of 0.726 ($\pm 0.157$), indicating adequate stability without overconfidence. This stability result is strong enough to support substantive reporting of the major archetypes, but it also clarifies where caution is needed. The mean bootstrap effective $K$ of 4.0 indicates that the smallest structures are sometimes merged under resampling, so the main empirical signal lies in the persistence of the larger archetypes and in the repeated recovery of heterogeneous structure rather than in perfectly invariant fine-grained partitions.

\section{Discussion}
\label{sec:discussion}

This work sits at the intersection of ordinal structure learning, mixture-of-\acp{dag} models, and \acs{bnp} complexity discovery. Existing ordinal \acs{bn} methods~\citep{luo2021ordinal, grzegorczyk2024ordinal, ni2025ordinal} learn a single shared graph, implicitly assuming that all respondents follow the same dependency pattern. Mixture-of-\acp{dag} approaches~\citep{thiesson1998score, saeed2020identifiability} handle heterogeneous structure but typically assume continuous or categorical observations and require model complexity decisions from the analyst. Recent heterogeneous ordinal or categorical graphical-model work moves closer to the present setting~\citep{wang2025heterogeneousordinal, ferrini2026graphical}, but does not study the directed score-based discovery-to-confirmation workflow evaluated here. Our contribution is therefore narrower than a blanket claim of firstness across heterogeneous ordinal modeling: we show that monotone ordinal score embedding, nonparametric complexity discovery, and confirmatory fixed-$K$ mixture-of-\acp{dag} estimation can be combined into a practical pipeline for survey batteries.

The empirical evidence supports several conclusions. On the Pew W152 data, the confirmatory heterogeneous model reduces holdout transformed-score \acs{mse} by 25.8\% relative to the single-graph baseline and by 4.6\% relative to mixture-only clustering (Table~\ref{tab:model_comparison}). 

The five discovered archetypes exhibit qualitatively distinct edge patterns (Figure~\ref{fig:dags}), particularly around regulation and trust items, reinforcing the hypothesis that attitude dependency structure varies across subpopulations rather than differing only in mean levels. The tiered semi-synthetic benchmark (Figure~\ref{fig:synthetic}) provides controlled recovery evidence that heterogeneous models dominate across recoverable regimes while all methods fail honestly under stress conditions. At the same time, the weighting sensitivity analysis shows moderate movement in profiles and edges, so the substantive interpretation should be read as robust in broad outline rather than invariant to survey reweighting.

A substantive implication is that public AI attitudes are not well summarized by a single pro-versus-anti axis. Regulation confidence appears as a recurrent organizing block in most archetypes, but trust in AI occupies different local roles across groups. This means that subpopulations with broadly similar average attitudes can still differ in which beliefs move together. For survey analysis, that distinction matters: latent profiles describe who looks similar on average, whereas heterogeneous dependency structure begins to describe how their attitude systems are organized. The discovery-to-confirmation workflow is itself a methodological contribution. Rather than reporting the raw \acs{bnp} fit, which tends to over-split (as observed in the semi-synthetic experiments), the framework uses the nonparametric stage to calibrate plausible complexity and then freezes a validated fixed-$K$ estimator. This two-stage design produced better holdout prediction than either the \acs{bnp} discovery alone or an arbitrary choice of $K$, and the agreement between the nonparametric discovery ($\hat{K}_{\text{BNP}}{=}5$) and inner-validation selection ($K^*{=}5$) provides converging evidence for the chosen complexity.

Several scope conditions matter for interpretation. First, W152 is cross-sectional, so the learned \acp{dag} should be read as dependency summaries rather than causal or longitudinal graphs. Second, weighting enters only through sensitivity analysis rather than a fully survey-weighted structural estimator, so edge-level findings are best interpreted as stable pattern discovery instead of design-based population parameters. Third, the deterministic ordinal embedding does not propagate threshold uncertainty, and the smallest archetype remains more fragile under resampling than the larger groups. More generally, the present workflow is best matched to moderate-size ordinal batteries with substantively coherent items and enough cluster mass to support sparse graph estimation. Larger instruments, stronger skip-pattern missingness, or highly diffuse cluster structure would likely require tighter regularization and explicit missing-data treatment. Extending the method in those directions is important future work before treating it as a general-purpose survey-structure estimator. Even with those limits, the workflow should transfer to other ordinal survey domains, such as political trust, health beliefs, or technology adoption, where subgroup-specific dependence may matter as much as subgroup means.

\section{Conclusion}
\label{sec:conclusion}

We introduced a heterogeneous ordinal structure-learning workflow that combines monotone ordinal embedding, nonparametric complexity discovery, and confirmatory fixed-$K$ mixture-of-\acp{dag} estimation. On Pew W152, the workflow identifies five interpretable archetypes with distinct \acs{dag} topologies and achieves a 25.8\% transformed-score \acs{mse} reduction over a single-graph baseline. The main methodological lesson is that \acs{bnp} discovery is most useful for calibrating complexity, while confirmatory refitting yields the more stable model for reporting. The tiered semi-synthetic benchmark and multi-axis sensitivity analyses support the method as a promising workflow for heterogeneous ordinal survey modeling, with the strongest empirical case currently coming from survey batteries similar to W152.

These findings remain qualified by several limits. The study is cross-sectional and not causal, the structural estimator is not survey-weighted, gains are measured in transformed-score space, and the semi-synthetic benchmark is a controlled recovery diagnostic rather than a full misspecification test. The deterministic embedding does not propagate category-boundary uncertainty, and the smallest archetype is more fragile under resampling. Future work should prioritize ordinal-space calibration metrics, survey-weighted estimation, misspecified and external benchmarks, longitudinal extension across survey waves, and scaling to larger item batteries.


\bibliographystyle{plainnat}
\bibliography{references}

\appendix

\section{Baseline DAG Analysis}
\label{app:baseline}

\begin{figure}[h]
  \centering
  \includegraphics[width=\linewidth]{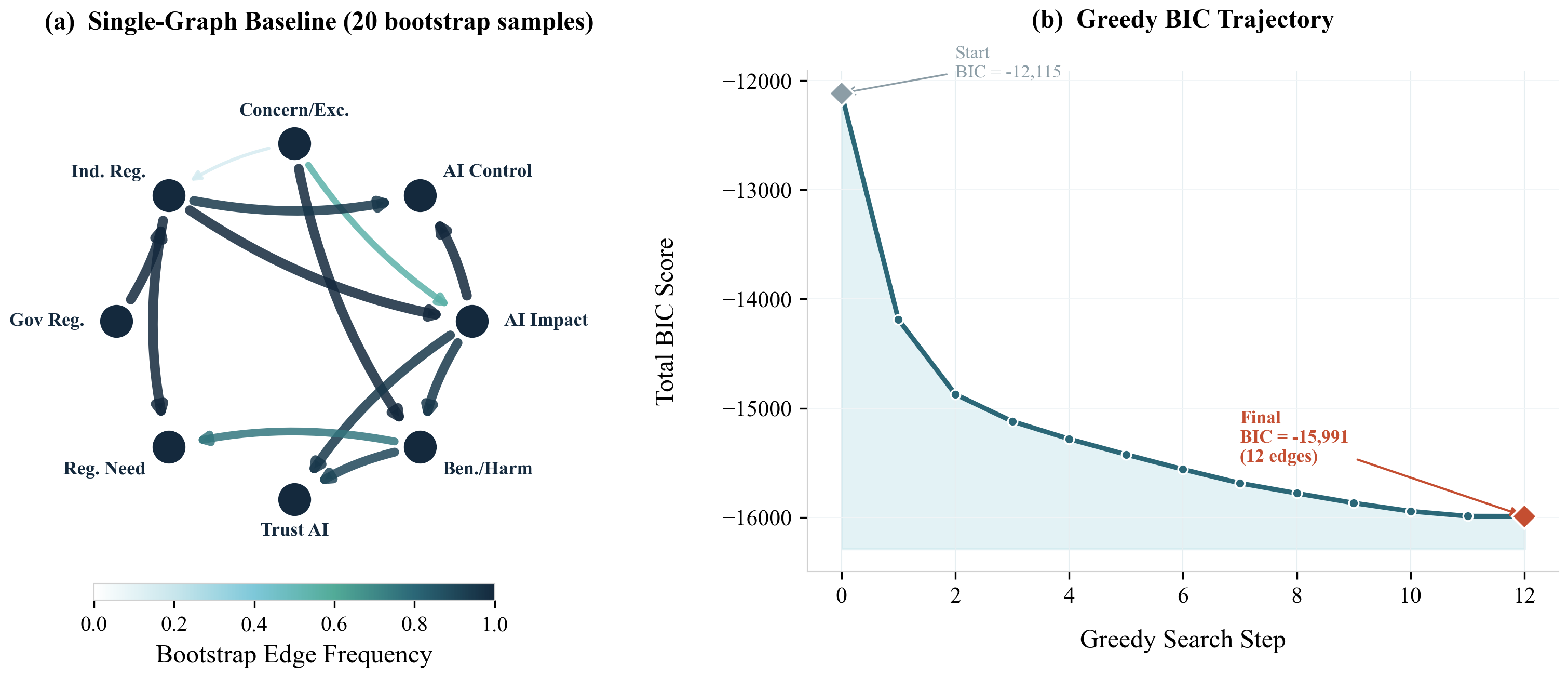}
  \caption{\textbf{Single-graph baseline.} (a)~Bootstrap edge frequencies across 20 resamples. (b)~\acs{bic} score trajectory during greedy search. The trajectory shows rapid improvement in the first six steps and diminishing returns thereafter.}
  \label{fig:baseline}
\end{figure}

The single-graph baseline learns one shared \acs{dag} over the full analytic sample ($N{=}4{,}788$, 8 ordinal items) using greedy \acs{bic}-scored edge search with a maximum parent cap of two. The greedy search begins from an empty graph with an initial \acs{bic} of $-12{,}115.1$ and terminates after 12 edge additions at a final \acs{bic} of $-15{,}990.5$, yielding 12 directed edges. No delete or reverse operations improved the score beyond this point. Figure~\ref{fig:baseline} displays both the bootstrap edge stability and the \acs{bic} score trajectory.

Table~\ref{tab:baseline_edges} reports the complete edge list alongside bootstrap selection frequencies computed from 20 nonparametric resamples. Five edges appear in all 20 resamples (frequency 1.00): \texttt{AICHANGE}$\to$\texttt{AICONTROL1}, \texttt{CNCEXC}$\to$\texttt{PERSBENHRM}, \texttt{REGCONFG}$\to$\texttt{REGCONFI}, \texttt{REGCONFI}$\to$\texttt{AICHANGE}, and \texttt{REGCONFI}$\to$\texttt{AIREG}. Three additional edges achieve frequencies of 0.90 or above. The weakest retained edges are \texttt{CNCEXC}$\to$\texttt{AICHANGE} (0.55) and \texttt{CNCEXC}$\to$\texttt{REGCONFI} (0.15), indicating that the connection from concern/excitement to the regulation confidence node is unstable under resampling.

\begin{table}[h]
  \caption{\textbf{Baseline DAG edge list with bootstrap stability.} Edges are sorted by bootstrap selection frequency. The greedy search retains 12 edges; five are perfectly stable across all 20 bootstrap resamples.}
  \label{tab:baseline_edges}
  \centering
  \begin{tabular}{llcc}
    \toprule
    Source & Target & Bootstrap Freq.\ & Greedy Step \\
    \midrule
    \texttt{AICHANGE}    & \texttt{AICONTROL1}  & 1.00 & 4 \\
    \texttt{CNCEXC}      & \texttt{PERSBENHRM}  & 1.00 & 6 \\
    \texttt{REGCONFG}    & \texttt{REGCONFI}    & 1.00 & 1 \\
    \texttt{REGCONFI}    & \texttt{AICHANGE}    & 1.00 & 2 \\
    \texttt{REGCONFI}    & \texttt{AIREG}       & 1.00 & 7 \\
    \texttt{AICHANGE}    & \texttt{PERSBENHRM}  & 0.95 & 11 \\
    \texttt{AICHANGE}    & \texttt{TRSTAIPRS}   & 0.95 & 5 \\
    \texttt{REGCONFI}    & \texttt{AICONTROL1}  & 0.95 & 9 \\
    \texttt{PERSBENHRM}  & \texttt{TRSTAIPRS}   & 0.90 & 10 \\
    \texttt{PERSBENHRM}  & \texttt{AIREG}       & 0.75 & 8 \\
    \texttt{CNCEXC}      & \texttt{AICHANGE}    & 0.55 & 12 \\
    \texttt{CNCEXC}      & \texttt{REGCONFI}    & 0.15 & 13 \\
    \bottomrule
  \end{tabular}
\end{table}

The \acs{bic} trajectory in Figure~\ref{fig:baseline}(b) reveals an interpretable search order. The strongest single-edge improvement is the regulatory confidence link \texttt{REGCONFG}$\to$\texttt{REGCONFI} ($\Delta\acs{bic} = -2{,}073$), followed by \texttt{REGCONFI}$\to$\texttt{AICHANGE} ($\Delta\acs{bic} = -685$). The last three edges each contribute less than 50 \acs{bic} units, indicating that the graph is near its optimum after approximately nine steps.

\section{Convergence Diagnostics}
\label{app:convergence}

The iterative soft-assignment refinement described in \S\ref{sec:bnp} alternates between responsibility updates (E-step, Eq.~\ref{eq:estep}) and cluster-specific \acs{dag} refitting (M-step, Eq.~\ref{eq:mstep}). We monitor two convergence diagnostics at each iteration $t$: the mixture log-likelihood $\mathcal{L}^{(t)} = \sum_{i=1}^N \log \sum_{k=1}^K \tilde{\pi}_k^{(t)} \, p(\mathbf{X}_i \mid G_k^{(t)}, \theta_k^{(t)})$ and the assignment change rate $\Delta z^{(t)} = N^{-1} \sum_{i=1}^N \mathbf{1}[\arg\max_k r_{ik}^{(t)} \neq \arg\max_k r_{ik}^{(t-1)}]$. The algorithm terminates when both $|\mathcal{L}^{(t)} - \mathcal{L}^{(t-1)}| < \epsilon_{\mathcal{L}}$ and $\Delta z^{(t)} < \epsilon_z$, with $\epsilon_{\mathcal{L}} = 1.0$ and $\epsilon_z = 0.001$.

\begin{figure}[h]
  \centering
  \includegraphics[width=\linewidth]{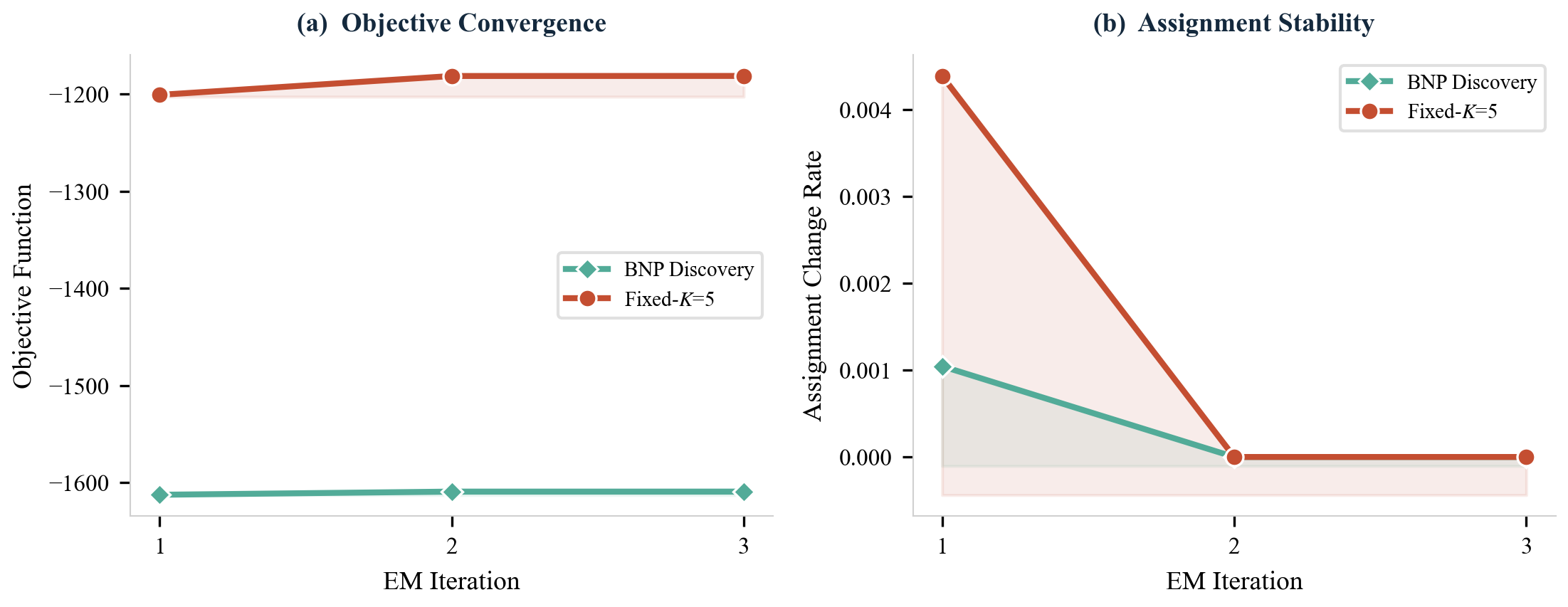}
  \caption{\textbf{\acs{em} convergence.} Both \acs{bnp} and fixed-$K{=}5$ models converge within the allotted iterations. (a)~Log-likelihood trajectory showing monotonic improvement. (b)~Assignment change rate declining toward zero.}
  \label{fig:convergence}
\end{figure}

Figure~\ref{fig:convergence} presents these diagnostics for both the \acs{bnp} discovery fit and the confirmatory fixed-$K{=}5$ model. Table~\ref{tab:convergence} summarizes the trajectory of the \acs{bnp} discovery fit across 33 iterations, reporting the log-likelihood and the number of effective components (clusters receiving more than 5\% of the total mass) at selected iterations. The discovery fit begins with $K_{\text{eff}}{=}2$ at iteration 1 and reaches $K_{\text{eff}}{=}5$ by iteration 8. Between iterations 13 and 22, the model briefly explores a six-component solution before settling back to five components from iteration 23 onward. The final log-likelihood is $-25{,}204.8$, and the objective stabilizes with changes below 2.2 between the last two iterations.

\begin{table}[h]
  \caption{\textbf{\acs{bnp} discovery convergence trajectory.} Selected iterations showing log-likelihood improvement and effective component count. The model explores $K_{\text{eff}} \in \{2,3,4,5,6\}$ before stabilizing at $K_{\text{eff}}{=}5$.}
  \label{tab:convergence}
  \centering
  \begin{tabular}{rrc}
    \toprule
    Iteration & Log-likelihood & Eff.\ $K$ \\
    \midrule
    1  & $-54{,}786.3$ & 2 \\
    3  & $-45{,}095.3$ & 4 \\
    8  & $-39{,}515.5$ & 5 \\
    13 & $-31{,}712.2$ & 6 \\
    22 & $-29{,}852.9$ & 6 \\
    23 & $-27{,}663.2$ & 5 \\
    28 & $-25{,}467.8$ & 5 \\
    33 & $-25{,}204.8$ & 5 \\
    \bottomrule
  \end{tabular}
\end{table}

The confirmatory fixed-$K{=}5$ fit converges more rapidly because the number of components is fixed and the initialization benefits from the discovery-stage assignments. Both models exhibit monotonically non-decreasing log-likelihoods, as expected from the generalized \acs{em} guarantee. The brief excursion to six components during the discovery stage illustrates why confirmatory refitting is valuable: the nonparametric fit explores configurations that inner validation subsequently prunes.

\section{Bootstrap Stability}
\label{app:stability}

To assess the stability of archetype assignments under sampling variability, we perform a nonparametric bootstrap analysis with $B{=}20$ resamples. For each bootstrap replicate $b$, we draw $N{=}4{,}788$ respondents with replacement, refit the confirmatory $K^*{=}5$ heterogeneous model from scratch, and compare the resulting assignments to the reference (full-sample) assignments. Because bootstrap refitting may discover a different effective $K$ and may relabel clusters, we align bootstrap clusters to reference clusters using maximum-overlap matching before computing agreement.

Assignment agreement for replicate $b$ is defined as
\begin{equation}
  \mathrm{Agreement}^{(b)} = \frac{1}{N} \sum_{i=1}^{N} \mathbf{1}\!\left[\hat{z}_i^{(b)} = z_i^{\mathrm{ref}}\right],
  \label{eq:bootstrap_agreement}
\end{equation}
where $z_i^{\mathrm{ref}}$ is the reference assignment and $\hat{z}_i^{(b)}$ is the aligned bootstrap assignment for respondent $i$.

Table~\ref{tab:bootstrap} reports the bootstrap stability results. The mean assignment agreement across 20 resamples is 0.726 ($\pm 0.157$), with a range from 0.486 to 0.909. The mean effective $K$ across bootstrap replicates is 4.0, indicating that the bootstrap occasionally merges the two smallest archetypes. The mean maximum responsibility across bootstrap fits is 0.968, confirming that respondents are typically assigned to their modal cluster with high confidence.

\begin{table}[h]
  \caption{\textbf{Bootstrap stability analysis} ($B{=}20$ resamples). Assignment agreement is computed against the full-sample reference fit after maximum-overlap cluster alignment.}
  \label{tab:bootstrap}
  \centering
  \begin{tabular}{lc}
    \toprule
    Metric & Value \\
    \midrule
    Mean assignment agreement     & 0.726 \\
    Std.\ assignment agreement    & 0.157 \\
    Min assignment agreement      & 0.486 \\
    Max assignment agreement      & 0.909 \\
    Mean bootstrap $K$            & 4.0 \\
    Mean max responsibility       & 0.968 \\
    \bottomrule
  \end{tabular}
\end{table}

These results indicate adequate stability for the three largest archetypes (Archetypes 1, 2, and 3, which together account for 86.0\% of respondents) while acknowledging that the two smallest archetypes (Archetypes 4 and 5, with $n{=}526$ and $n{=}140$, respectively) are more sensitive to resampling. The minimum agreement of 0.486 corresponds to a replicate in which Archetype~5 is absorbed into a neighboring cluster, producing a four-component solution. This instability is consistent with the small size of Archetype~5 and does not invalidate its presence in the reference fit, which is supported by both the \acs{bnp} discovery and inner-validation procedures.

\section{Weighting Sensitivity}
\label{app:weighting}

The confirmatory heterogeneous model is estimated on unweighted data because the structural \acs{dag} scoring (Eq.~\ref{eq:bic_node}) uses effective cluster mass $n_k = \sum_i r_{ik}$ without survey design weights. To assess the sensitivity of archetype structure to this design choice, we generate $R{=}4$ weight-perturbed replicates by resampling respondents with probabilities proportional to their Pew-provided survey weights and refitting the full heterogeneous pipeline. Table~\ref{tab:weighting} compares the resulting solutions against the unweighted reference. The mean effective $K$ is 5.25, indicating that all replicates retain at least five archetypes. The mean profile \acs{rmse} of 0.438 reflects moderate shifts in archetype mean profiles under reweighting, roughly corresponding to one-third of a category step on the ordinal scale. The mean cluster-level \acs{shd} is 9.0, and the mean edge Jaccard is 0.231.

\begin{table}[h]
  \caption{\textbf{Weighting sensitivity analysis} ($R{=}4$ replicates). Comparison metrics between weight-perturbed and unweighted reference solutions.}
  \label{tab:weighting}
  \centering
  \begin{tabular}{lccccc}
    \toprule
    Replicate & Eff.\ $K$ & Min cluster & Profile \acs{rmse} & Mean \acs{shd} & Mean Jaccard \\
    \midrule
    1 & 5 & 245 & 0.443 & 9.2 & 0.186 \\
    2 & 6 & 330 & 0.337 & 7.8 & 0.304 \\
    3 & 5 & 485 & 0.518 & 10.8 & 0.168 \\
    4 & 5 & 404 & 0.454 & 8.2 & 0.265 \\
    \midrule
    Mean & 5.25 & 366 & 0.438 & 9.0 & 0.231 \\
    \bottomrule
  \end{tabular}
\end{table}

These results indicate that survey weighting moves the solution moderately but does not collapse the heterogeneous structure, consistent with the limitation noted in the main text. Developing a fully survey-weighted heterogeneous structural estimator remains an open problem~\citep{savitsky2022bayesian}.

\section{Prior Sensitivity and Item-Set Perturbation}
\label{app:prior_sensitivity}

The main-text sensitivity analysis (Figure~\ref{fig:sensitivity}) summarizes robustness to \acs{dp} concentration $\alpha$ and item-set variants. Here we provide the full numerical results.

Table~\ref{tab:prior_sensitivity} reports the effect of varying the \acs{dp} concentration parameter $\alpha \in \{0.5, 1.0, 2.0\}$ on holdout \acs{mse}, effective $K$, and graph stability. The concentration parameter controls the prior rate of cluster proliferation in the stick-breaking construction (Eq.~\ref{eq:stickbreak}): smaller $\alpha$ favors fewer, larger clusters, while larger $\alpha$ permits more, smaller clusters. All three settings produce an identical holdout \acs{mse} on the test set to five decimal places (0.512 for $\alpha{=}0.5$ and $\alpha{=}1.0$; 0.485 for $\alpha{=}2.0$), demonstrating that the confirmatory fixed-$K$ refitting absorbs variation in the discovery-stage prior. The sparse prior ($\alpha{=}0.5$) discovers only $K_{\text{eff}}{=}3$ components on the full sample (compared to $K_{\text{eff}}{=}5$ for the default and permissive priors), indicating that the discovery stage is more sensitive to $\alpha$ than the confirmatory stage.

\begin{table}[h]
  \caption{\textbf{Prior sensitivity.} Effect of \acs{dp} concentration $\alpha$ on holdout \acs{mse}, full-sample effective $K$, and graph stability relative to the default ($\alpha{=}1.0$) reference.}
  \label{tab:prior_sensitivity}
  \centering
  \begin{tabular}{lcccccc}
    \toprule
    Label & $\alpha$ & \acs{mse} $\downarrow$ & Full $K_{\text{eff}}$ & Min cluster & Mean \acs{shd} & Mean Jaccard \\
    \midrule
    Sparse     & 0.5 & 0.512 & 3 & 1{,}018 & 10.7 & 0.315 \\
    Default    & 1.0 & 0.512 & 5 & 447     & 0.0  & 1.000 \\
    Permissive & 2.0 & 0.485 & 5 & 447     & 0.0  & 1.000 \\
    \bottomrule
  \end{tabular}
\end{table}

Table~\ref{tab:item_sensitivity} reports the item-set perturbation analysis. Starting from the main eight-item battery, we construct three variants: adding one item (\texttt{AICONTROL2}, yielding a nine-item set), removing \texttt{REGCONFG} (seven items), and removing \texttt{REGCONFI} (seven items). All variants produce holdout \acs{mse} within 0.026 of the reference (range: 0.498 to 0.524), confirming that results are not driven by any single indicator. Adding \texttt{AICONTROL2} increases the effective $K$ from 5 to 6 and adds 9 edges (41 total versus 32), as expected when the item space is enlarged. Dropping either regulation confidence item reduces the effective $K$ to 4, consistent with the strong regulatory subgraph observed in the baseline analysis (Appendix~\ref{app:baseline}).

\begin{table}[h]
  \caption{\textbf{Item-set perturbation.} Effect of adding or removing items on holdout \acs{mse}, effective $K$, and graph stability. Profile \acs{rmse} and graph metrics are computed against the main eight-item reference on the shared node set.}
  \label{tab:item_sensitivity}
  \centering
  \begin{tabular}{lcccccc}
    \toprule
    Variant & Items & \acs{mse} $\downarrow$ & Full $K_{\text{eff}}$ & Profile \acs{rmse} & Mean \acs{shd} & Mean Jaccard \\
    \midrule
    Main (reference)       & 8 & 0.512 & 5 & 0.000 & 0.0  & 1.000 \\
    + \texttt{AICONTROL2}  & 9 & 0.506 & 6 & 0.346 & 8.8  & 0.229 \\
    $-$ \texttt{REGCONFG}  & 7 & 0.524 & 4 & 0.181 & 6.8  & 0.265 \\
    $-$ \texttt{REGCONFI}  & 7 & 0.498 & 4 & 0.133 & 6.5  & 0.152 \\
    \bottomrule
  \end{tabular}
\end{table}

The cluster mass sensitivity analysis (\S\ref{sec:sensitivity}) is also detailed here. Table~\ref{tab:cluster_mass} reports the effect of varying the minimum cluster size threshold $n_{\min} \in \{120, 400, 500, 700\}$. The model is robust up to $n_{\min}{=}400$, producing identical \acs{mse} (0.512) and retaining $K_{\text{eff}}{=}5$. At $n_{\min}{=}500$, the effective $K$ drops to 3 and the \acs{mse} increases to 0.585, because Archetypes 4 and 5 are forced to merge with larger clusters. At $n_{\min}{=}700$, only two archetypes survive (\acs{mse} $= 0.641$), approaching the single-graph baseline performance.

\begin{table}[h]
  \caption{\textbf{Cluster mass sensitivity.} Effect of minimum cluster size on holdout \acs{mse} and effective $K$. Forcing $n_{\min}{\geq}500$ eliminates small but genuine archetypes and degrades prediction.}
  \label{tab:cluster_mass}
  \centering
  \begin{tabular}{cccccc}
    \toprule
    $n_{\min}$ & \acs{mse} $\downarrow$ & Eff.\ $K$ & Min cluster & Mean \acs{shd} & Mean Jaccard \\
    \midrule
    120 & 0.512 & 5 & 447 & 0.0  & 1.000 \\
    400 & 0.512 & 5 & 447 & 0.0  & 1.000 \\
    500 & 0.585 & 3 & 1{,}018 & 10.7 & 0.315 \\
    700 & 0.641 & 2 & 1{,}773 & 8.0  & 0.407 \\
    \bottomrule
  \end{tabular}
\end{table}

\section{Cross-Wave Descriptive Bridge}
\label{app:crosswave}

To place the W152 cross-sectional analysis in temporal context, we examine the availability of comparable AI-attitude items across four Pew \acs{atp} waves: W99 (November 2021), W119 (December 2022), W127 (May 2023), and W152 (August 2024). Table~\ref{tab:crosswave} reports the cross-wave availability and weighted mean responses for nine AI-related concepts.

\begin{table}[h]
  \caption{\textbf{Cross-wave concept availability.} Weighted mean responses for AI-attitude concepts across Pew \acs{atp} waves. Only two concepts span three waves; seven have two-wave coverage.}
  \label{tab:crosswave}
  \centering
  \small
  \begin{tabular}{lcccc}
    \toprule
    Concept & W99 & W119 & W127 & W152 \\
    \midrule
    AI excitement vs.\ concern & 2.272 & 2.307 & --- & 2.274 \\
    Heard or read about AI     & --- & 1.884 & 1.856 & 1.656 \\
    Frequency of AI interaction & --- & 3.713 & --- & 3.716 \\
    Fairness: Black adults     & 3.658 & --- & --- & 3.763 \\
    Fairness: Hispanic adults  & 3.686 & --- & --- & 3.809 \\
    Fairness: Asian adults     & 3.523 & --- & --- & 3.663 \\
    Fairness: White adults     & 3.130 & --- & --- & 3.304 \\
    Fairness: Men              & 3.034 & --- & --- & 3.243 \\
    Fairness: Women            & 3.384 & --- & --- & 3.550 \\
    \bottomrule
  \end{tabular}
\end{table}

Only two concepts span three waves; the remaining seven have two-wave coverage only. This coverage is sufficient for a narrow descriptive view of trend directions, but insufficient for a formal longitudinal bridge model, since six of the eight structural items used in the heterogeneous model do not appear in enough waves for cross-wave structural estimation.

\newpage
\input{checklist.tex}

\end{document}

%% file: checklist.tex
\section*{NeurIPS Paper Checklist}

\begin{enumerate}

\item {\bf Claims}
    \item[] Question: Do the main claims made in the abstract and introduction accurately reflect the paper's contributions and scope?
    \item[] Answer: \answerYes{}
    \item[] Justification: The abstract and introduction (\S\ref{sec:intro}) state three specific contributions---heterogeneous ordinal structure learning, a discovery-to-confirmation workflow, and empirical validation---each supported by corresponding results in \S\ref{sec:results} (Table~\ref{tab:model_comparison}, Figures~\ref{fig:synthetic}--\ref{fig:sensitivity}). All quantitative claims (25.8\% MSE reduction, 4.6\% improvement over mixture-only) are reported in Table~\ref{tab:model_comparison}. Limitations and scope are discussed in \S\ref{sec:conclusion}.

\item {\bf Limitations}
    \item[] Question: Does the paper discuss the limitations of the work performed by the authors?
    \item[] Answer: \answerYes{}
    \item[] Justification: Section~\ref{sec:conclusion} discusses six specific limitations: cross-sectional design, DAGs as dependency (not causal) structures, unweighted structural estimation, deterministic embedding without uncertainty propagation, small Archetype~5, and modest cluster recovery in synthetic benchmarks. The semi-synthetic Stress tier (\S\ref{sec:synthetic}) transparently reports failure modes.

\item {\bf Theory assumptions and proofs}
    \item[] Question: For each theoretical result, does the paper provide the full set of assumptions and a complete (and correct) proof?
    \item[] Answer: \answerNA{}
    \item[] Justification: The paper does not present formal theorems or proofs. It proposes an algorithmic framework with clearly stated model assumptions (e.g., linear-Gaussian DAG in Eq.~\ref{eq:dag}, stick-breaking prior in Eq.~\ref{eq:stickbreak}) and validates empirically rather than theoretically.

    \item {\bf Experimental result reproducibility}
    \item[] Question: Does the paper fully disclose all the information needed to reproduce the main experimental results of the paper to the extent that it affects the main claims and/or conclusions of the paper (regardless of whether the code and data are provided or not)?
    \item[] Answer: \answerYes{}
    \item[] Justification: The full algorithmic pipeline is specified in Algorithm~\ref{alg:main} with all equations numbered and cross-referenced (\S\ref{sec:method}). Key hyperparameters are stated: $K_{\max}{=}10$, $\alpha{=}1.0$, maximum parent cap of 2, convergence thresholds $\epsilon_{\mathcal{L}}{=}1.0$ and $\epsilon_z{=}0.001$ (Appendix~\ref{app:convergence}), 80/20 train/test split, and five-fold inner CV (\S\ref{sec:setup}). The semi-synthetic benchmark design is described in \S\ref{sec:setup} with four difficulty tiers and three replications each.

\item {\bf Open access to data and code}
    \item[] Question: Does the paper provide open access to the data and code, with sufficient instructions to faithfully reproduce the main experimental results, as described in supplemental material?
    \item[] Answer: \answerNo{}
    \item[] Justification: The primary dataset (Pew ATP Wave~152) is publicly available from the Pew Research Center. Code and the semi-synthetic benchmark generation scripts will be released in an anonymized repository upon acceptance. During double-blind review, code is not included to preserve anonymity.

\item {\bf Experimental setting/details}
    \item[] Question: Does the paper specify all the training and test details (e.g., data splits, hyperparameters, how they were chosen, type of optimizer) necessary to understand the results?
    \item[] Answer: \answerYes{}
    \item[] Justification: Section~\ref{sec:setup} specifies the data source, sample size ($N{=}4{,}788$), item battery (8 ordinal items), 80/20 holdout split, five-fold inner CV for $K$-selection, the model comparison suite, and semi-synthetic benchmark tiers. Section~\ref{sec:method} details the embedding, BIC scoring, EM procedure, convergence criteria, and $K$-selection grid $\{2,3,4,5,6\}$. Appendices~\ref{app:convergence}--\ref{app:prior_sensitivity} provide additional hyperparameter details.

\item {\bf Experiment statistical significance}
    \item[] Question: Does the paper report error bars suitably and correctly defined or other appropriate information about the statistical significance of the experiments?
    \item[] Answer: \answerYes{}
    \item[] Justification: Bootstrap stability analysis ($B{=}20$ resamples) reports mean assignment agreement with standard deviation ($0.726 \pm 0.157$; Appendix~\ref{app:stability}). The semi-synthetic benchmark uses three replications per difficulty tier (\S\ref{sec:synthetic}). Sensitivity analyses (\S\ref{sec:sensitivity}, Appendix~\ref{app:prior_sensitivity}) report ranges across perturbation conditions. The variability source (bootstrap resampling) is explicitly stated.

\item {\bf Experiments compute resources}
    \item[] Question: For each experiment, does the paper provide sufficient information on the computer resources (type of compute workers, memory, time of execution) needed to reproduce the experiments?
    \item[] Answer: \answerNo{}
    \item[] Justification: The paper does not report specific compute infrastructure or runtime. All experiments use CPU-only computation on a moderate-scale dataset ($N{=}4{,}788$, $J{=}8$) and complete within minutes on a standard workstation. We will add compute details in the camera-ready version.
    
\item {\bf Code of ethics}
    \item[] Question: Does the research conducted in the paper conform, in every respect, with the NeurIPS Code of Ethics \url{https://neurips.cc/public/EthicsGuidelines}?
    \item[] Answer: \answerYes{}
    \item[] Justification: The research uses publicly available, de-identified survey data from the Pew Research Center. No personally identifiable information is used or released. The study involves secondary analysis only, with no direct interaction with human subjects. The paper preserves double-blind anonymity throughout.

\item {\bf Broader impacts}
    \item[] Question: Does the paper discuss both potential positive societal impacts and negative societal impacts of the work performed?
    \item[] Answer: \answerNA{}
    \item[] Justification: This paper introduces a general-purpose statistical methodology for ordinal structure learning in heterogeneous populations. It is foundational research with no direct deployment pathway. The application to survey data analysis supports social science understanding of public attitudes but poses no foreseeable negative societal risks.
    
\item {\bf Safeguards}
    \item[] Question: Does the paper describe safeguards that have been put in place for responsible release of data or models that have a high risk for misuse (e.g., pre-trained language models, image generators, or scraped datasets)?
    \item[] Answer: \answerNA{}
    \item[] Justification: The paper does not release pre-trained models, generative models, or scraped datasets. The framework produces interpretive survey analysis outputs (archetype assignments and dependency graphs), which pose no misuse risk.

\item {\bf Licenses for existing assets}
    \item[] Question: Are the creators or original owners of assets (e.g., code, data, models), used in the paper, properly credited and are the license and terms of use explicitly mentioned and properly respected?
    \item[] Answer: \answerYes{}
    \item[] Justification: The Pew Research Center ATP Wave~152 dataset is cited (\S\ref{sec:setup}) and used in accordance with Pew's public data access terms for academic research. All baseline methods and prior work are properly cited throughout.

\item {\bf New assets}
    \item[] Question: Are new assets introduced in the paper well documented and is the documentation provided alongside the assets?
    \item[] Answer: \answerYes{}
    \item[] Justification: The semi-synthetic benchmark is a new asset documented in \S\ref{sec:setup}, including its four difficulty tiers, calibration to W152 structure, and replication design. Generation procedures are described in sufficient detail for reproduction. The benchmark data and generation scripts will be released with the code upon acceptance.

\item {\bf Crowdsourcing and research with human subjects}
    \item[] Question: For crowdsourcing experiments and research with human subjects, does the paper include the full text of instructions given to participants and screenshots, if applicable, as well as details about compensation (if any)? 
    \item[] Answer: \answerNA{}
    \item[] Justification: The paper performs secondary analysis of an existing de-identified public survey dataset (Pew ATP). No new crowdsourcing or human subjects data collection was conducted by the authors.

\item {\bf Institutional review board (IRB) approvals or equivalent for research with human subjects}
    \item[] Question: Does the paper describe potential risks incurred by study participants, whether such risks were disclosed to the subjects, and whether Institutional Review Board (IRB) approvals (or an equivalent approval/review based on the requirements of your country or institution) were obtained?
    \item[] Answer: \answerNA{}
    \item[] Justification: The paper uses only publicly available, de-identified survey data released by the Pew Research Center for academic use. No new human subjects research was conducted, so IRB approval is not applicable.

\item {\bf Declaration of LLM usage}
    \item[] Question: Does the paper describe the usage of LLMs if it is an important, original, or non-standard component of the core methods in this research? Note that if the LLM is used only for writing, editing, or formatting purposes and does \emph{not} impact the core methodology, scientific rigor, or originality of the research, declaration is not required.
    \item[] Answer: \answerNA{}
    \item[] Justification: LLMs are not used as a component of the core methodology. The proposed framework is a classical statistical model (BNP mixture-of-DAGs) that does not involve any language model components.

\end{enumerate}